\documentclass[11pt,a4paper]{article}
\usepackage[hyperref]{aacl-ijcnlp2020}
\usepackage{times}
\usepackage{latexsym}

\usepackage{url}
\usepackage{enumitem}
\usepackage{subfig}
\usepackage{booktabs}
\usepackage{makecell}
\usepackage{graphicx}
\usepackage{pgfplots}
\pgfplotsset{width=6.4cm,compat=1.9}
\usepackage{multirow}
\usepackage{tabularx}

\usepackage{microtype}

\aclfinalcopy 
\def\aclpaperid{22} 

\title{Intent Detection with WikiHow}

\author{Li Zhang \\\And
  Qing Lyu \\
  University of Pennsylvania \\
  {\tt \{zharry,lyuqing,ccb\}@seas.upenn.edu} \\\And
  Chris Callison-Burch \\}

\date{}

\begin{document}
\maketitle
\begin{abstract}
Modern task-oriented dialog systems need to reliably understand users' intents. Intent detection is even more challenging when moving to new domains or new languages, since there is little annotated data. To address this challenge, we present a suite of pretrained intent detection models which can predict a broad range of intended goals from many actions because they are trained on wikiHow, a comprehensive instructional website. Our models achieve state-of-the-art results on the Snips dataset, the Schema-Guided Dialogue dataset, and all 3 languages of the Facebook multilingual dialog datasets. Our models also demonstrate strong zero- and few-shot performance, reaching over 75\% accuracy using only 100 training examples in all datasets.\footnote{The data and models are available at \url{https://github.com/zharry29/wikihow-intent}.}
\end{abstract}

\section{Introduction}
Task-oriented dialog systems like Apple's Siri, Amazon Alexa, and Google Assistant have become pervasive in smartphones and smart speakers. To support a wide range of functions, dialog systems must be able to map a user's natural language instruction onto the desired skill or API. Performing this mapping is called intent detection.

Intent detection is usually formulated as a sentence classification task. Given an utterance (e.g. ``wake me up at 8''), a system needs to predict its intent (e.g. ``Set an Alarm''). Most modern approaches use neural networks to jointly model intent detection and slot filling \cite{6707709,liu2016attentionbased,goo-etal-2018-slot,zhang-etal-2019-joint}. In response to a rapidly growing range of services, more attention has been given to zero-shot intent detection \cite{ferreira2015zero,ferreira15,yazdani-henderson-2015-model,chen2016zero,kumar2017zero,gangadharaiah-narayanaswamy-2019-joint}. While most existing research on intent detection proposed novel model architectures, few have attempted data augmentation. One such work \cite{10.1145/1526709.1526773} showed that models can learn much knowledge that is important for intent detection from massive online resources such as Wikipedia.

We propose a pretraining task based on wikiHow, a comprehensive instructional website with over 110,000 professionally edited articles. Their topics span from common sense such as ``How to Download Music'' to more niche tasks like ``How to Crochet a Teddy Bear.'' We observe that the header of each step in a wikiHow article describes an action and can be approximated as an utterance, while the title describes a goal and can be seen as an intent. For example, ``find good gas prices'' in the article ``How to Save Money on Gas'' is similar to the utterance ``where can I find cheap gas?'' with the intent ``Save Money on Gas.'' Hence, we introduce a dataset based on wikiHow, where a model predicts the goal of an action given some candidates. Although most of wikiHow's domains are far beyond the scope of any present dialog system, models pretrained on our dataset would be robust to emerging services and scenarios. Also, as wikiHow is available in 18 languages, our pretraining task can be readily extended to multilingual settings.

Using our pretraining task, we fine-tune transformer language models, achieving state-of-the-art results on the intent detection task of the Snips dataset \cite{coucke2018snips}, the Schema-Guided Dialog (SGD) dataset \cite{rastogi2019towards}, and all 3 languages (English, Spanish, and Thai) of the Facebook multilingual dialog datasets \cite{schuster-etal-2019-cross-lingual}, with statistically significant improvements. As our accuracy is close to 100\% on all these datasets, we further experiment with zero- or few-shot settings. Our models achieve over 70\% accuracy with no in-domain training data on Snips and SGD, and over 75\% with only 100 training examples on all datasets. This highlights our models' ability to quickly adapt to new utterances and intents in unseen domains. 

\section{WikiHow Pretraining Task}

\subsection{Corpus}
We crawl the wikiHow website in English, Spanish, and Thai (the languages were chosen to match those in  the Facebook multilingual dialog datasets). 
We define the \textbf{goal} of each artcle  as its title stripped of the  prefix ``How to'' (and its equivalent in other languages).  We extract a set of \textbf{steps} for each article, by taking the bolded header of each paragraph.

\subsection{WikiHow Pretraining Dataset}
A wikiHow article's goal can approximate an intent, and each step in it can approximate an associated utterance. We formulate the pretraining task as a 4-choose-1 multiple choice format: given a step, the model infers the correct goal among 4 candidates. For example, given the step ``let check-in agents and flight attendants know if it’s a special occasion'' and the candidate goals:\\
\hspace*{15pt} A. Get Upgraded to Business Class\\
\hspace*{15pt} B. Change a Flight Reservation \\
\hspace*{15pt} C. Check Flight Reservations \\ 
\hspace*{15pt} D. Use a Discount Airline Broker\\
the correct goal would be A.  This is similar to intent detection, where a system is given a user utterance and then must select a supported intent.

We create intent detection pretraining data using goal-step pairs from each wikiHow article.  Each article contributes at least one positive goal-step pair.  However, it is challenging to sample negative candidate goals for a given step. There are two reasons for this.  First, random sampling of goals correctly results in true negatives, but they tend to be so distant from the positive goal that the classification task becomes trivial and the model does not learn sufficiently.  Second, if we sample goals that are similar to the positive goal, then they might not be true negatives, since there are many steps in wikiHow often with overlapping goals. To sample high-quality negative training instances, we start with the correct goal and search in its article's ``related articles'' section for an article whose title has the least lexical overlap with the current goal. We recursively do this until we have enough candidates. Empirically, examples created this way are mostly clean, with an example shown above. We select one positive goal-step pair from each article by picking its longest step. 
In total, our wikiHow pretraining datasets have 107,298 English examples,  64,803 Spanish examples, and  6,342 Thai examples. 

\section{Experiments}
We fine-tune a suite of off-the-shelf language models pretrained on our wikiHow data, and evaluate them on 3 major intent detection benchmarks.

\vspace{-.25cm}
\subsection{Models}
We fine-tune a pretrained RoBERTa model \cite{DBLP:journals/corr/abs-1907-11692} for the English datasets and a pretrained XLM-RoBERTa model \cite{conneau2019unsupervised} for the multilingual datasets. We cast the instances of the intent detection datasets into a multiple-choice format, where the utterance is the input and the full set of intents are the possible candidates, consistent with our wikiHow pretraining task. For each model, we append a linear classification layer with cross-entropy loss to calculate a likelihood for each candidate, and output the candidate with the maximum likelihood. 

For each intent detection dataset in any language, we consider the following settings:\\
\textbf{+in-domain} (+ID): a model is only trained on the dataset's in-domain training data;\\
\textbf{+wikiHow +in-domain} (+WH+ID): a model is first trained on our wikiHow data in the corresponding language, and then trained on the dataset's in-domain training data;\\
\textbf{+wikiHow zero-shot} (+WH 0-shot): a  model is trained only on our wikiHow data in the corresponding language, and then applied directly to the dataset's evaluation data.

For non-English languages, the corresponding wikiHow data might suffer from smaller sizes and lower quality. Hence, we additionally consider the following cross-lingual transfer settings for non-English datasets:\\
\textbf{+en wikiHow +in-domain} (+enWH+ID), a model is trained on wikiHow data in English, before it is trained on the dataset's in-domain training data;\\
\textbf{+en wikiHow zero-shot} (+enWH 0-shot), a model is trained on wikiHow data in English, before it is directly applied to the dataset's evaluation data.\\

\begin{table}[t!]
\fontsize{10}{11}\selectfont
\centering
\begin{tabularx}{\columnwidth}{lrrrr}
\toprule
 & \makecell{Training\\Size} & \makecell{Valid.\\Size} & \makecell{Test\\Size} & \makecell{Num.\\Intents} \\ \midrule
Snips & 2,100 & 700 & N/A & 7  \\
SGD & 163,197 & 24,320 & 42,922 & 4  \\ 
FB-en & 30,521 & 4,181 & 8,621 & 12  \\
FB-es & 3,617 & 1,983 & 3,043 & 12  \\ 
FB-th & 2,156 & 1,235 & 1,692 & 12  \\ 
\bottomrule
\end{tabularx}
\caption{Statistics of the dialog benchmark datasets.\vspace{-.3cm}}
\label{table:datasets}
\end{table}

\vspace{-.3cm}
\subsection{Datasets}

We consider the 3 following benchmarks: \\
\textbf{The Snips dataset} \cite{coucke2018snips} is a single-turn English dataset. It is one of the most cited dialog benchmarks in recent years, containing utterances collected from the Snips personal voice assistant. While its full training data has 13,784 examples, we find that our models only need its smaller training split consisting of 2,100 examples to achieve high performance. Since Snips does not provide test sets, we use the validation set for testing and the full training set for validation. Snips involves 7 intents, including \textit{Add to Playlist}, \textit{Rate Book}, \textit{Book Restaurant}, \textit{Get Weather}, \textit{Play Music}, \textit{Search Creative Work}, and \textit{Search Screening Event}. Some example utterances include ``Play the newest melody on Last Fm by Eddie Vinson,'' ``Find the movie schedule in the area,'' etc.\\
\textbf{The Schema-Guided Dialogue dataset} (SGD) \cite{rastogi2019towards} is a multi-turn English dataset. It is the largest dialog corpus to date spanning dozens of domains and services, used in the DSTC8 challenge \cite{rastogi2020schemaguided} with dozens of team submissions. Schemas are provided with at most 4 intents per dialog turn. Examples of these intents include \textit{Buy Movie Tickets for a Particular show}, \textit{Make a Reservation with the Therapist}, \textit{Book an Appointment at a Hair Stylist}, \textit{Browse attractions in a given city}, etc. At each turn, we use the last 3 utterances as input. An example: ``That sounds fun. What other attractions do you recommend? There is a famous place of worship called Akshardham.''\\
\textbf{The Facebook multilingual datasets} (FB-en/es/th) \cite{schuster-etal-2019-cross-lingual} is a single-turn multilingual dataset. It is the only multilingual dialog dataset to the best of our knowledge, containing utterances annotated with intents and slots in English (en), Spanish (es), and Thai (th). It involves 12 intents, including \textit{Set Reminder}, \textit{Check Sunrise}, \textit{Show Alarms}, \textit{Check Sunset}, \textit{Cancel Reminder}, \textit{Show Reminders}, \textit{Check Time Left on Alarm}, \textit{Modify Alarm}, \textit{Cancel Alarm}, \textit{Find Weather}, \textit{Set Alarm}, and \textit{Snooze Alarm}. Some example utterances are ``Is my alarm set for 10 am today?'' ``Colocar una alarma para mañana a las 3 am,'' \includegraphics[height=14pt]{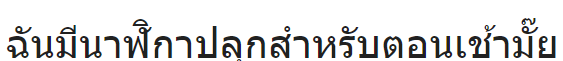}  etc.
\\
Statistics of the datasets are shown in Table~\ref{table:datasets}.

\subsection{Baselines}
We compare our models with the previous state-of-the-art results of each dataset: \\
\textbullet\ \citet{9082602}
proposed a Siamese neural
network with triplet loss, achieving state-of-the-art results on Snips and FB-en; \\
\textbullet\ \citet{8907842} used multi-task learning to jointly learn intent detection and slot filling, achieving state-of-the-art results on FB-es and FB-th;  \\
\textbullet\ \citet{ma2019endtoend} augmented the data via back-translation to and from Chinese, achieving state-of-the-art results on SGD.

\begin{table}[t!]
\fontsize{10}{11}\selectfont
\begin{tabularx}{\columnwidth}{lXXl}
\toprule
          & Snips & SGD & FB-en  \\ \midrule
\cite{9082602}      & .993  & N/A & .993 \\
\cite{ma2019endtoend}      & N/A  & .948 & N/A \\ \midrule
+in-domain (+ID)  & .990  & .942 & .993 \\
(ours) +WH+ID   & \textbf{.994}  & \textbf{.951$\dagger$} & \textbf{.995$\dagger$} \\
(ours) +WH 0-shot & .713  & .787 & .445 \\ \midrule
Chance    & .143  & .250 & .083 \\
\bottomrule
\end{tabularx}
\caption{The accuracy of intent detection on English datasets using RoBERTa. State-of-the-art performances are in bold; $\dagger$ indicates statistically significant improvement from the previous state-of-the-art.}
\label{table:eng-results}
\end{table}

\begin{table}[t!]
\fontsize{10}{11}\selectfont
\begin{tabularx}{\columnwidth}{lXXXX}
\toprule
          & FB-en & FB-es & FB-th  \\ \midrule
\cite{9082602} & .993  & N/A & N/A \\
\cite{8907842} &  N/A  & .978 & .967 \\ \midrule
+in-domain (+ID) & .993  & .986 & .962 \\
(ours) +WH+ID   & \textbf{.995}  & .988 & .971 \\
(ours) +enWH+ID   & \textbf{.995}  & \textbf{.990$\dagger$} & \textbf{.976$\dagger$} \\
(ours) +WH 0-shot & .416  & .129 & .119 \\
(ours) +enWH 0-shot & .416  & .288 & .124 \\ \midrule
Chance    & .083  & .083 & .083 \\
\bottomrule
\end{tabularx}
\caption{The accuracy of intent detection on multilingual datasets using XLM-RoBERTa.  \vspace{-.3cm} }
\label{table:multi-results}
\end{table}

\begin{figure*}\small
\centering
\begin{tikzpicture}
\begin{axis}[
    title={Snips (RoBERTa)},
    axis x line*=bottom,
    axis y line*=left,
    legend style={nodes={scale=0.7, transform shape}},
    xmode=log,
    log ticks with fixed point,
    ymin=0, ymax=1.05,
    ytick={0,.2,.4,.6,.8,1},
    x filter/.code=\pgfmathparse{#1 + 6.90775527898214},
    legend pos=north west,
    xticklabels={,,}
]
\addplot[mark=triangle,densely dotted,mark options={solid}] 
table {
0.01	0.117
0.05	0.137
0.1	0.47
0.5	0.84
1	0.963
}; 
\addplot[mark=square*,mark options={scale=0.8}] 
table {
0.01	0.823
0.05	0.9
0.1	0.953
0.5	0.974
1	.994
}; 
\draw ({rel axis cs:0,0}|-{axis cs:0,0.143}) -- ({rel axis cs:1,0}|-{axis cs:0,0.143});
\addplot[
    only marks,
    visualization depends on=\thisrow{alignment} \as \alignment,
    nodes near coords,point meta=explicit symbolic,every node near coord/.style={anchor=\alignment}
    ] table [
     meta index=2
     ] {
x       y       label       alignment
0.1 0.953 .953 -90
0.1 0.47 .470 150
};
\end{axis}
\end{tikzpicture}%
\hskip 1pt 
\begin{tikzpicture}
\begin{axis}[
    title={SGD (RoBERTa)},
    axis x line*=bottom,
    axis y line*=left,
    legend style={nodes={scale=0.7, transform shape}},
    xmode=log,
    ymin=0, ymax=1.05,
    ytick={0,.2,.4,.6,.8,1},
    log ticks with fixed point,
    x filter/.code=\pgfmathparse{#1 + 6.90775527898214},
    legend pos=north west,
    xticklabels={,,},
    yticklabels={,,}
]
\addplot[mark=triangle,densely dotted,mark options={solid}] 
table {
0.01	0.441
0.05	0.493
0.1	0.531
0.5	0.586
1	0.817
}; 
\addplot[mark=square*,mark options={scale=0.8}] 
table {
0.01	0.629
0.05	0.694
0.1	0.755
0.5	0.829
1	.860
}; 
\draw ({rel axis cs:0,0}|-{axis cs:0,0.250}) -- ({rel axis cs:1,0}|-{axis cs:0,0.250});
\addplot[
    only marks,
    visualization depends on=\thisrow{alignment} \as \alignment,
    nodes near coords,point meta=explicit symbolic,every node near coord/.style={anchor=\alignment}
    ] table [
     meta index=2
     ] {
x       y       label       alignment
0.1 0.531 .531 90
0.1 0.755 .755 -90
};
\end{axis}
\end{tikzpicture}%
\hskip 1pt 
\begin{tikzpicture}
\begin{axis}[
    title={FB-en (RoBERTa)},
    axis x line*=bottom,
    axis y line*=left,
    legend style={nodes={scale=0.7, transform shape}},
    xmode=log,
    ymin=0, ymax=1.05,
    ytick={0,.2,.4,.6,.8,1},
    log ticks with fixed point,
    x filter/.code=\pgfmathparse{#1 + 6.90775527898214},
    legend pos=north west,
    xticklabels={,,},
    yticklabels={,,}
]
\addplot[mark=triangle,densely dotted,mark options={solid}] 
table {
0.01	0.383
0.05	0.458
0.1	0.458
0.5	0.911
1	0.981
}; 
\addplot[mark=square*,mark options={scale=0.8}] 
table {
0.01	0.359
0.05	0.8
0.1	0.884
0.5	0.962
1	0.982
}; 
\draw ({rel axis cs:0,0}|-{axis cs:0,0.083}) -- ({rel axis cs:1,0}|-{axis cs:0,0.083});
\addplot[
    only marks,
    visualization depends on=\thisrow{alignment} \as \alignment,
    nodes near coords,point meta=explicit symbolic,every node near coord/.style={anchor=\alignment}
    ] table [
     meta index=2
     ] {
x       y       label       alignment
0.1 0.458 .458 150
0.1 0.884 .884 -40
};
\end{axis}
\end{tikzpicture}%
\hskip 1pt 
\begin{tikzpicture}
\begin{axis}[
    title={FB-en (XLM-RoBERTa)},
    axis x line*=bottom,
    axis y line*=left,
    legend style={nodes={scale=0.7, transform shape}},
    xmode=log,
    ymin=0, ymax=1.05,
    ytick={0,.2,.4,.6,.8,1},
    log ticks with fixed point,
    x filter/.code=\pgfmathparse{#1 + 6.90775527898214},
    legend pos=north west,
]
\addplot[mark=triangle,densely dotted,mark options={solid}] 
table {
0.01	0.405
0.05	0.459
0.1	0.481
0.5	0.75
1	0.951
}; 
\addplot[mark=square*,mark options={scale=0.8}] 
table {
0.01	0.667
0.05	0.789
0.1	0.894
0.5	0.964
1	0.98
}; 
\draw ({rel axis cs:0,0}|-{axis cs:0,0.083}) -- ({rel axis cs:1,0}|-{axis cs:0,0.083});
\addplot[
    only marks,
    visualization depends on=\thisrow{alignment} \as \alignment,
    nodes near coords,point meta=explicit symbolic,every node near coord/.style={anchor=\alignment}
    ] table [
     meta index=2
     ] {
x       y       label       alignment
0.1 0.894 .894 -40
0.1 0.481 .481 150
};
\end{axis}
\end{tikzpicture}%
\hskip 1pt 
\begin{tikzpicture}
\begin{axis}[
    title={FB-es (XLM-RoBERTa)},
    axis x line*=bottom,
    axis y line*=left,
    legend style={nodes={scale=0.7, transform shape}},
    xmode=log,
    ymin=0, ymax=1.05,
    ytick={0,.2,.4,.6,.8,1},
    log ticks with fixed point,
    x filter/.code=\pgfmathparse{#1 + 6.90775527898214},
    legend pos=north west,
    yticklabels={,,},
]
\addplot[mark=triangle,densely dotted,mark options={solid}] 
table {
0.01	0.29
0.05	0.138
0.1	0.349
0.5	0.646
1	0.902
}; 
\addplot[mark=square*,mark options={scale=0.8}] 
table {
0.01	0.143
0.05	0.244
0.1	0.633
0.5	0.804
1	0.925
}; 
\addplot[mark=o,densely dashed,mark options={solid}] 
table {
0.01	0.405
0.05	0.638
0.1	0.845
0.5	0.945
1	0.978
}; 
\draw ({rel axis cs:0,0}|-{axis cs:0,0.083}) -- ({rel axis cs:1,0}|-{axis cs:0,0.083});
\addplot[
    only marks,
    visualization depends on=\thisrow{alignment} \as \alignment,
    nodes near coords,point meta=explicit symbolic,every node near coord/.style={anchor=\alignment}
    ] table [
     meta index=2
     ] {
x       y       label       alignment
0.1 0.845 .845 -50
0.1 0.663 .663 -100
0.1 0.349 .349 150
};
\end{axis}
\end{tikzpicture}%
\hskip 1pt 
\begin{tikzpicture}
\begin{axis}[
    title={FB-th (XLM-RoBERTa)},
    axis x line*=bottom,
    axis y line*=left,
    legend style={nodes={scale=0.7, transform shape}},
    xmode=log,
    ymin=0, ymax=1.05,
    ytick={0,.2,.4,.6,.8,1},
    log ticks with fixed point,
    x filter/.code=\pgfmathparse{#1 + 6.90775527898214},
    legend style={at={(0.52,0.02)},anchor=south west},
    yticklabels={,,},
]
\addplot[mark=triangle,densely dotted,mark options={solid}] 
table {
0.01 .275
0.05 .161
0.1  .341
0.5  .517 
1    .881
}; \addlegendentry{+ID}
\addplot[mark=square*,mark options={scale=0.8}] 
table {
0.01	0.216
0.05	0.602
0.1	0.851
0.5	0.912
1	0.966
}; \addlegendentry{(ours)+WH+ID}
\addplot[mark=o,densely dashed,mark options={solid}] 
table {
0.01	0.151
0.05	0.665
0.1	0.853
0.5	0.919
1	0.97
}; \addlegendentry{(ours)+enWH+ID}
\draw ({rel axis cs:0,0}|-{axis cs:0,0.083}) -- ({rel axis cs:1,0}|-{axis cs:0,0.083});
\addlegendimage{}
\addlegendentry{Chance}
\addplot[
    only marks,
    visualization depends on=\thisrow{alignment} \as \alignment,
    nodes near coords,point meta=explicit symbolic,every node near coord/.style={anchor=\alignment}
    ] table [
     meta index=2
     ] {
x       y       label       alignment
0.1 0.853 .853 -40
0.1 0.851 .851 150
0.1 0.341 .341 -40
};
\end{axis}
\end{tikzpicture}
\caption{Learning curves of models in low-resource settings. The vertical axis is the accuracy of intent detection, while the horizontal axis is the number of in-domain training examples of each task, distorted to log-scale.\vspace{-.5cm}}
\label{fig:learning}
\end{figure*}
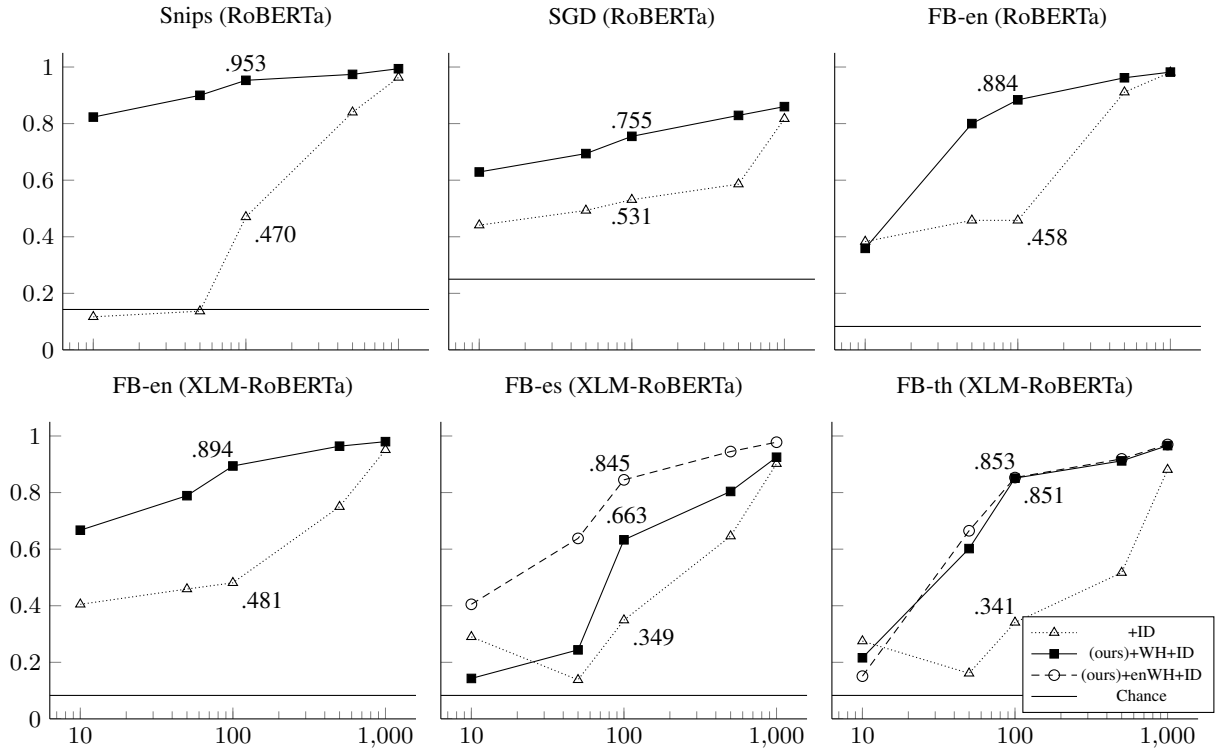

\subsection{Modelling Details}
After experimenting with base and large models, we use RoBERTa-large for the English datasets and XLM-RoBERTa-base for the multilingual dataset for best performances. All our models are implemented using the HuggingFace Transformer library\footnote{\url{https://github.com/huggingface/transformers}}. 

We tune our model hyperparameters on the validation sets of the datasets we experiment with. However, in all cases, we use a unified setting which empirically performs well, using the Adam optimizer \cite{kingma2014adam} with an epsilon of $1e^{-8}$, a learning rate of $5e^{-6}$, maximum sequence length of 80 and 3 epochs. We variate the batch size from 2 to 16 according to the number of candidates in the multiple-choice task, to avoid running out of memory. We save the model every 1,000 training steps, and choose the model with the highest validation performance to be evaluated on the test set. 

We run our experiments on an NVIDIA GeForce RTX 2080 Ti GPU, with half-precision floating point format (FP16) with O1 optimization. Each epoch takes up to 90 minutes in the most resource intensive setting, i.e. running a RoBERTa-large on around 100,000 training examples of our wikiHow pretraining dataset. 

\subsection{Results}

The performance of RoBERTa on the English datasets (Snips, SGD, and FB-en) are shown in Table~\ref{table:eng-results}. We repeat each experiment 20 times, report the mean accuracy, and calculate its p-value against the previous state-of-the-art result, using a one-sample and one-tailed t-test with a significance level of 0.05. Our models achieve state-of-the-art results using the available in-domain training data. Moreover, our wikiHow data enables our models to demonstrate strong performances in zero-shot settings with no in-domain training data, implying our models' strong potential to adapt to new domains. 

The performance of XLM-RoBERTa on the multilingual datasets (FB-en, FB-es, and FB-th) are shown in Table~\ref{table:multi-results}. Our models achieve state-of-the-art results on all 3 languages. While our wikiHow data in Spanish and Thai does improve models' performances, its effect is less salient than the English wikiHow data.

Our experiments above focus on settings where all available in-domain training data are used. However, modern task-oriented dialog systems must rapidly adapt to burgeoning services (e.g. Alexa Skills) in different languages, where little training data are available. To simulate low-resource settings, we repeat the experiments with exponentially increasing number of training examples up to 1,000. We consider the models trained only on in-domain data (+ID), those first pretrained on our wikiHow data in corresponding languages (+WH+ID), and those first pretrained on our English wikiHow data (+enWH+ID) for FB-es and FB-th. 

The learning curves of each dataset are shown in Figure~\ref{fig:learning}. Though the vanilla transformers models (+ID) achieve close to state-of-the-art performance with access to the full training data (see Table~\ref{table:eng-results} and \ref{table:multi-results}), they struggle in the low-resource settings. When given up to 100 in-domain training examples, their accuracies are less than 50\% on most datasets. In contrast, our models pretrained on our wikiHow data (+WH+ID) can reach over 75\% accuracy given only 100 training examples on all datasets. 

\vspace{-.1cm}
\section{Discussion and Future Work}
As our model performances exceed 99\% on Snips and FB-en, the concern arises that these intent detection datasets are ``solved''. We address this by performing error analysis and providing future outlooks for intent detection. 

\subsection{Error Analysis}
Our model misclassifies 7 instances in the Snips test set. Among them, 6 utterances include proper nouns on which intent classification is contingent. For example, the utterance ``please open Zvooq'' assumes the knowledge that Zvooq is a streaming service, and its labelled intent is ``Play Music.'' 

Our model misclassifies 43 instances in the FB-en test set. Among them, 10 has incorrect labels: e.g. the labelled intent of ``have alarm go off at 5 pm'' is ``Show Alarms,'' while our model prediction ``Set Alarm'' is in fact correct. 28 are ambiguous: e.g. the labelled intent of ``repeat alarm every weekday'' is ``Set Alarm,'' whereas that of ``add an alarm for 2:45 on every Monday'' is ``Modify Alarm.'' We only find 1 example an interesting edge case: the gold intent of ``remind me if there will be a rain forecast tomorrow'' is ``Find Weather,'' while our model incorrectly chooses ``Set Reminder.''

By performing manual error analyses on our model predictions, we observe that most misclassified examples involve ambiguous wordings, wrong labels, or obscure proper nouns. Our observations imply that Snips and FB-en might be too easy to effectively evaluate future models. 

\subsection{Open-Domain Intent Detection}

State-of-the-art models now achieve greater than 99\% percent accuracy on standard benchmarks for intent detection. 
However, intent detection is far from being solved. The standard benchmarks only have a dozen intents, but future dialog systems will need to support many more functions with intents from a wide range of domains. To demonstrate that our pretrained models can adapt to unseen, open-domain intents, we hold out 5,000 steps (as utterances) with their corresponding goals (as intents) from our wikiHow dataset as a proxy of an intent detection dataset with more than 100,000 possible intents (all goals in wikiHow). 

For each step, we sample 100 goals with the highest embedding similarity to the correct goal, as most other goals are irrelevant. We then rank them for the likelihood that the step helps achieve them. Our RoBERTa model achieves a mean reciprocal rank of 0.462 and a 36\% accuracy of ranking the correct goal first. As a qualitative example, given the step ``find the order that you want to cancel,'' the top 3 ranked steps are ``Cancel an Order on eBay'', ``Cancel an Online Order'', ``Cancel an Order on Amazon.'' This hints that our pretrained models' can work with a much wider range of intents than those in current benchmarks, and suggests that future intent detection research should focus on open domains, especially those with little data.

\vspace{-.15cm}
\section{Conclusion}
By pretraining language models on wikiHow, we attain state-of-the-art results in 5 major intent detection datasets spanning 3 languages. The wide-ranging domains and languages of our pretraining resource enable our models to excel with few labelled examples in multilingual settings, and suggest open-domain intent detection is now feasible. 

\section*{Acknowledgments}

This research is based upon work supported in part by the DARPA KAIROS Program (contract FA8750-19-2-1004), the DARPA LwLL Program (contract FA8750-19-2-0201), and the IARPA BETTER Program (contract 2019-19051600004). Approved for Public Release, Distribution Unlimited. The views and conclusions contained herein are those of the authors and should not be interpreted as necessarily representing the official policies, either expressed or implied, of DARPA, IARPA, or the U.S. Government. 

We thank the anonymous reviewers for their valuable feedback. 

\bibliography{aacl-ijcnlp2020}
\bibliographystyle{acl_natbib}

\appendix

\end{document}